\documentclass[conference,a4paper]{IEEEtran} 
\usepackage{graphicx}
\usepackage[T1]{fontenc}
\usepackage{upquote}
\usepackage{amsmath}
\usepackage{float}
\usepackage{placeins}
\usepackage{multicol}
\usepackage{multirow}
\usepackage{subfig}
\usepackage[noadjust]{cite}
\usepackage{float}
\usepackage{adjustbox}
\usepackage{amsmath}
\usepackage[linesnumbered,vlined,ruled]{algorithm2e}
\usepackage{enumitem}
\usepackage[nodisplayskipstretch]{setspace}
\begin{document}

\abovedisplayskip=0pt
\abovedisplayshortskip=-1pt
\belowdisplayskip=0pt
\belowdisplayshortskip=-1pt


\title{Realtime Multilevel Crowd Tracking using Reciprocal Velocity Obstacles}

\author{
\IEEEauthorblockN{Aniket Bera \& Dinesh Manocha}
University of North Carolina at Chapel Hill\\
\texttt{http://gamma.cs.unc.edu/RCrowdT}\\
}

\maketitle

\begin{abstract}

We present a novel, realtime algorithm to compute the trajectory of each pedestrian in moderately dense crowd 
scenes. Our formulation is based on an adaptive particle filtering scheme that uses a multi-agent motion model based on velocity-obstacles, and takes into account local interactions as well as physical and personal constraints of each pedestrian. Our method dynamically changes the number of particles allocated to each pedestrian based on different confidence metrics.
Additionally, we use a new high-definition crowd video dataset, which is used to evaluate the performance of different pedestrian tracking algorithms. This dataset consists of videos of indoor and outdoor scenes, recorded at different locations with 30-80 pedestrians. We highlight the performance benefits of our algorithm over prior techniques using this dataset. In practice, our algorithm can compute trajectories of tens of pedestrians on a multi-core desktop CPU at interactive rates (27-30 frames per second). To the best of our knowledge, our approach is 4-5 times faster than prior methods, which provide similar accuracy.
\end{abstract}


\section{INTRODUCTION}

Tracking pedestrians in a crowd is a well-studied problem in computer vision, robotics, and related areas. The goal is to spatially and temporally localize each moving pedestrian in a video and compute its trajectory. As autonomous robots and driverless cars are increasingly used in the physical world with tens or hundreds of pedestrians, it is important to \textit{track}, and also \textit{predict} motion and behavior at realtime rates. 
We also need real-time crowd tracking capabilities for surveillance activities~\cite{benfold2011stable}, evaluating crowd behaviors~\cite{shoval2006application}, detecting anomalous behavior~\cite{mehran2009abnormal}, crowd counting~\cite{sidla2006pedestrian}, realtime evacuation planning~\cite{doulamis2009evacuation}, collision-free navigation in dynamics scenes, etc. 

\noindent The problem of pedestrian tracking has been extensively studied and a variety of techniques have been proposed. 
In many ways, pedestrians correspond to the most difficult categories
of object tracking. Pedestrians tend to change their speed to avoid collisions with obstacles
and other pedestrians. 
Large variations in their appearance and illumination makes
it hard for color-based template tracking algorithms to continuously track a pedestrian. In crowded scenes,
the pairwise interactions between pedestrians can increase significantly and add to the complexity of predictive tracking schemes. 
Some of the most reliable tracking methods are developed for offline, non-realtime applications, where the knowledge of future frames is used and the algorithms make multiple passes over the video frames. Different approaches have been proposed for online or realtime pedestrian tracking, but they are currently limited to simple scenes with a few pedestrians (e.g. less than 10).  Moreover, a major challenge is to handle scenes with higher crowd density, i.e. when a high number of pedestrians are located in a small area (e.g. 3-4 pedestrians per squared meter).

\begin{figure}[ht]
	\centering
		\includegraphics[width=0.5\textwidth]{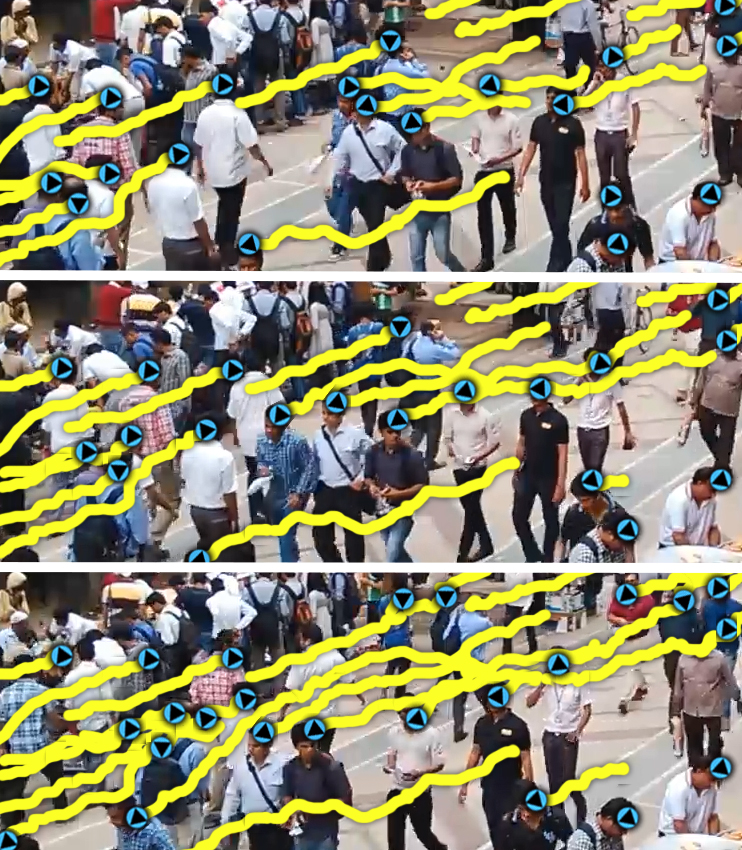}
\caption{{\em Tracking Street Crowds: Our algorithm achieves high accuracy in this dense street dataset with 144 pedestrians and can track at 27fps on a multi-core desktop CPU (Dataset - NPLACE-3).
}}
	\label{fig:1}
\end{figure}

\noindent In real-world scenarios, the trajectory of each pedestrian is governed by its intermediate goal location as well as local interactions with other pedestrians and obstacles in the scene. This includes avoiding collisions and computing an efficient path towards the goal. One of the major challenges in crowd tracking is to exploit these characteristics and use an appropriate motion model for each pedestrian. Some of the widely used motion models are based on constant velocity or constant acceleration~\cite{del2011particle}, though they may not work well in dense situations.


{\bf Main Results:} We present a novel realtime multilevel tracking algorithm for dense crowds that uses particle filters. Our approach dynamically changes the number of particles allocated to each pedestrian based on multiple confidence metrics. We use a non-linear parametric multi-agent motion model, Reciprocal Velocity Obstacles (RVO)~\cite{van2011reciprocal}, which takes into account reactive behavior of pedestrians in a dense setting and is used to compute the confidence metrics. Our approach aims
to significantly decrease the 
computational cost for realtime tracking and can easily be 
generalized to other multi-model particle filters. 

\begin{figure*}[thb]
\centering
\subfloat[Online Boosting]{\label{fig:img2}\includegraphics[width=0.31\linewidth]{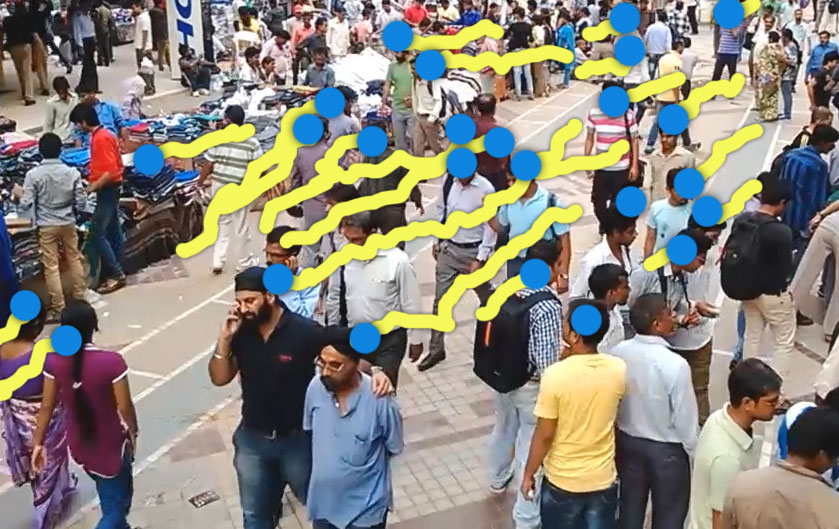}} \ 
\subfloat[Mean-shift]{\label{fig:img3}\includegraphics[width=0.31\linewidth]{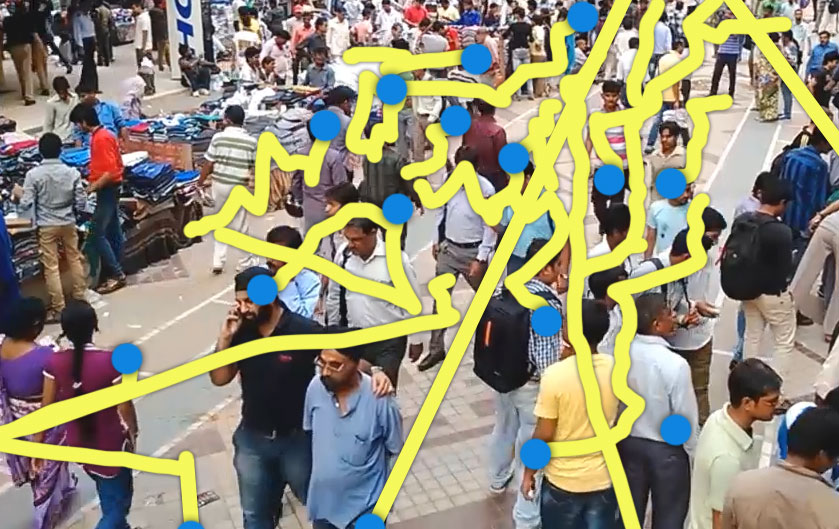}} \ 
\subfloat[Our Approach]{\label{fig:img4}\includegraphics[width=0.31\linewidth]{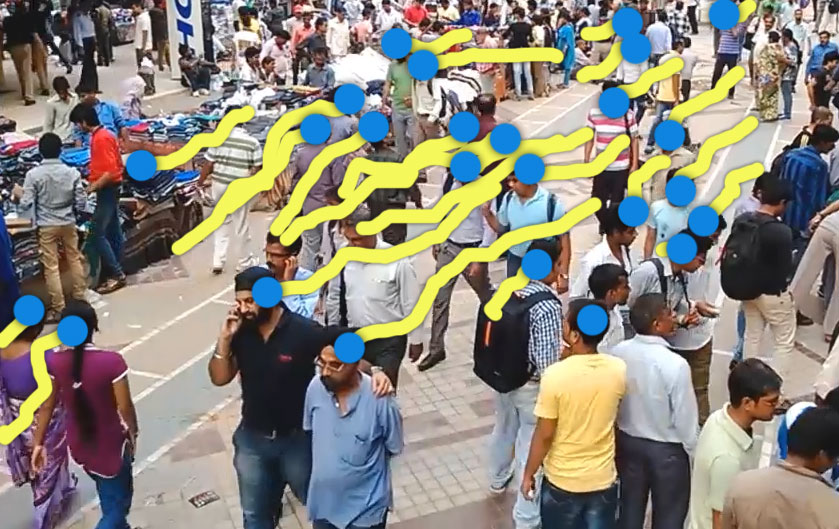}} 
\vspace*{-0.5em}
\caption{{\em Performance comparison of our approach with other algorithms on a crowded scene.:
(a) Online Boosting~\cite{babenko2009visual}(11 fps);
(b) MeanShift algorithm~\cite{cheng1995mean} (33 fps);
(c) Our Approach MLPF-RVO (28 fps). Overall, the accuracy of our approach is comparable to Online Boosting and the performance is a little slower than the MeanShift algorithm (Dataset - NPLACE-2)
}}
\label{fig:2}
\end{figure*}

\noindent We use RVOs to model the state transition distribution, which includes the motion prior pedestrian state.  We also estimate and iteratively refine the RVO parameters, which improves the accuracy of the motion model for tracking prediction and the
confidence measures. 

\noindent We evaluate the performance of our algorithm on new datasets, which include both indoor and outdoor scenes recorded at different locations with 30 - 80 pedestrians and compare with prior methods. Our algorithm can track tens of
pedestrians at realtime rates (i.e. more than 25fps) on a multi-core CPU, In practice, our approach is about 4-5 times faster than prior methods, that provide similar accuracy.

\section{Related Work}

\noindent In this section, we briefly review some prior work on pedestrian tracking. 
We also refer the reader to some surveys~\cite{enzweiler2009monocular,yilmaz2006object}.

\noindent Pedestrian tracking algorithms can be classified as either online or offline: online 
trackers use only the present or past frames, while offline trackers also use data from future frames. Zhang et al.~\cite{zhang2012real} proposed an online approach that uses non-adaptive random projections 
to model the structure of the image feature space of objects. Oron et al.~\cite{oron2012locally} 
described a method to estimate the amount of local deformation in rigid or deformable objects. The 
color-based probabilistic tracking method proposed  by Perez et al.~\cite{perez2002color} is fast
 but prone to loss of trajectories from occlusion. Collins's method~\cite{collins2003mean} tracked blob 
via mean-shifts and Jia et al.~\cite{jia2012visual} presented a method to track objects 
using a local sparse appearance model. Kwon et al.~\cite{kwon2011tracking} proposed a method which adaptively switches trackers and the trackers are sampled using the Markov Chain Monte
Carlo method from a predefined tracker space. Particle filters have been widely used for online tracking~\cite{okuma2004boosted}~\cite{khan2004mcmc}~\cite{nummiaro2002object}. Some of the state of the art accurate tracking methods are offline~\cite{sharma2012unsupervised}~\cite{rodriguez2011density}. However, some of these methods require future-state information, may make multiple passes over the video frames, and are not useful for realtime applications.

\noindent Many crowd tracking algorithms use motion models to 
improve tracking accuracy and prediction. Song et al.~\cite{song2013fully} proposed an approach to cluster the trajectories following the notion of ``persons only appear/disappear at entry/exit''. Ali et al.~\cite{ali2008floor} presented a method based on floor-fields to compute the probability of motion in highly dense crowded scenes. Kratz et al.~\cite{kratz2012going} and Zhao et al.~\cite{zhao2012tracking} presented an approach using local motion patterns in dense videos.
Rodriguez et al.~\cite{rodriguez2011data} used a large collection of public crowd videos to learn crowd motion patterns by extracting global video features. These methods are  well suited for dense crowds that can be characterized by a given motion pattern. Though the most commonly used motion model in pedestrian tracking are linear single-agent models, including constant velocity and constant acceleration~\cite{del2011particle}.
Other motion models used in pedestrian tracking are the Social Force model~\cite{helbing1995social}~\cite{bera2014}~\cite{mehran2009abnormal}, LTA~\cite{pellegrini2009you} and ATTR~\cite{yamaguchi2011you}.



\section{Our Approach}

\subsection{Algorithm}

\begin{figure}[h]
	\centering
\includegraphics[width=0.5\textwidth]{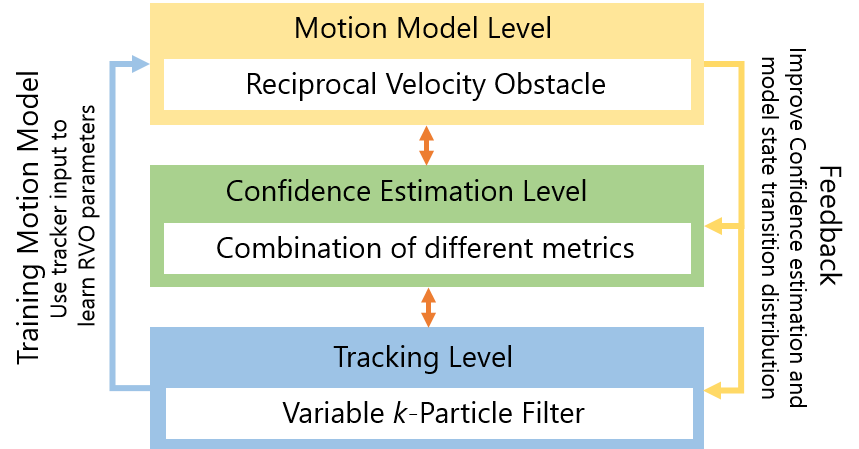}
\caption{{\em Our algorithm uses three levels to track each pedestrian in a crowd.  In the first level, we calculate the tracker output using a variable particle filter based approach. In the second level, we calculate the confidence of our tracker using a motion model centric metric approach. The number of particles used to track a pedestrian, k, vary over different frames based on the confidence estimate. In the third level, we estimate and iteratively refine the RVO-based motion parameters, which provides a continuous feedback loop to the other levels.  }}
	\label{fig:algo}
\end{figure}


In this section, we give an overview of approach and present details of our motion model. 
Our underlying tracking algorithm is based on particle filters. 
The particle filter is a parametric method which solves
non-Gaussian and non-linear state estimation problems~\cite{arulampalam2002tutorial}. Since
it can recover from lost tracks and occlusions, the
particle filters are frequently used in object tracking. However, its performance
can be compute intensive and the cost is directly proportional to the number of particles being used per pedestrian. However, with more particles, the probability of tracking a pedestrian accurately is higher and vice-versa. As a result, we need to use appropriate number of particles to balance the tradeoffs between computation cost and accuracy. Ideally, we would use lesser particles (lower $k$) at most times and increase $k$ only when needed, e.g. when there is a large change in motion trajectory, lighting, appearance or partial occlusions. We propose a {\em multi-level}(MLPF) approach that adaptively computes $k$ for every pedestrian at each timestep.

A particle filter-based tracker inherently depends on the use of a motion model, which propagates these particles to the next state. Some of the commonly used single-agent motion models are based on constant velocity or constant acceleration may not work in dense crowds. In these cases, the increased interactions between the pedestrians can breakdown the assumptions of constant velocity or constant acceleration.


\subsection{Reciprocal Velocity Obstacles}

\noindent RVO predicts all the pedestrians' states at every timestep given the state information from the current timestep~\cite{van2011reciprocal}. RVO is a local collision avoidance and navigation algorithm that enables the state of the agents to evolve into locally collision-avoidance states during the next time step. Each agent or pedestrian is represented as a 2D circle in the plane, and the state information for each agent consists of radius, current position and velocity, preferred velocity for the next timestep that is governed by the intermediate goal position. The RVO algorithm assumes that each pedestrian also knows the current position and velocity of other nearby agents. 

\noindent Let $\mathbf v_{pref}$ be the preferred velocity for a pedestrian that is  based on the intermediate goal location. The RVO formulation takes into account the position and velocity of each neighboring pedestrian to compute the new velocity. The velocity of the neighbors is used to formulate the ORCA constraints for local collision avoidance~\cite{van2011reciprocal}. The computation of new velocity is expressed as an
optimization problem for each pedestrian.
If an agent's preferred velocity is forbidden by the ORCA constraints, that agent chooses the closest velocity that lies in the feasible region: 

\begin{equation} \label{eqn:ORCA}
\mathbf{v}_{RVO} = \underset{\mathbf{v} \notin OCRA}{\arg\max} \|\mathbf{v} - \mathbf v_{pref}\|.
\end{equation}

\noindent The ORCA constraints are represented as the boundary of a half plane containing the space of all collision-free velocities. We highlight the computation for two pedestrians, say $x_{1}$ and $x_{2}$. The minimum vector $\mathbf{u}$ of the change in relative velocity to avoid a collision is computed. ORCA requires each pedestrian to change its current velocity by at least
$\frac{1}{2}\mathbf{u}$. Then the boundary of the ORCA constraint corresponds to a line containing the point $\mathbf{v}+\frac{1}{2}\vec{u}$  in the velocity space, with the direction perpendicular to $\mathbf{u}$. The ORCA constraint on $x_{1}$'s velocity induced by $x_{2}$ is given as: 
\begin{equation}\label{eqn:ORCA_X1X2}
ORCA_{x_{1}|x_{2}}=\{\mathbf{v}|(\mathbf{v}-(\mathbf{v}_{x_{1}}+\frac{1}{2} \mathbf{u}))\cdot \mathbf{\hat{u}} \geq 0\},
\end{equation}

\noindent where $\mathbf{v}_{x_{1}}$ is $x_{1}$'s current velocity and $\mathbf{\hat{u}}$ is the normalized vector $\mathbf{u}$.
More details and mathematical formulations of the ORCA constraints are given in~\cite{van2011reciprocal}.

\begin{figure}[h]
	\centering
\includegraphics[width=0.5\textwidth]{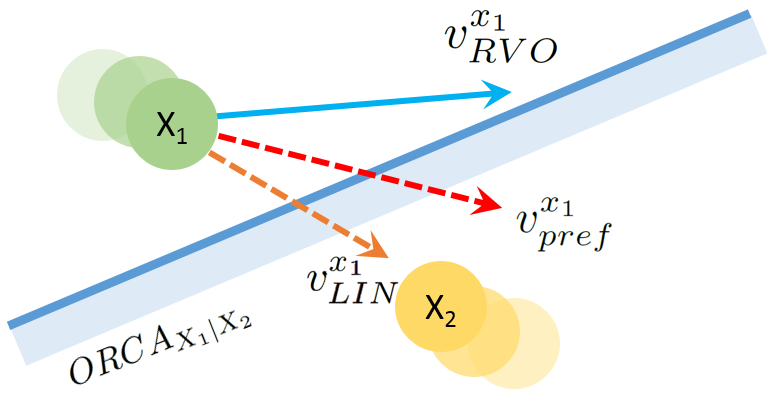}
\caption{{\em RVO Multi-agent Motion Model illustrating a pedestrian, $x_{1}$'s preferred velocity ($v_{pref}$) and the optimal collision free velocity computed by the RVO model ($v_{RVO}$). It is the velocity closest to $v_{pref}$, that lies in the feasible region. We also show the pedestrian velocity computed using the constant velocity model ($v_{LIN}$) which leads to a future collision state and hence, an incorrect prediction.}}
	\label{fig:5}
\end{figure}

\subsection{Multi-Level Particle Filter (MLPF)}

\noindent Our approach has three levels, as shown in Fig. 3. The first is the `tracking level', where we fit each pedestrian RVO state into the standard particle filter formulation. The second level is `confidence estimation', where we use multiple metrics to measure the reliability of the tracker and adaptively modify the number of active particles per pedestrian. The third level is the `motion model', where we estimate and iteratively refine the RVO parameters to best match our input video. We use this trained RVO model as an input to the levels below.

\textbf{Tracking Level:}
We use the standard particle filter and combine it with RVO parameters. 
Given a pedestrian's RVO state $x_{t}$ at time step $t$, RVO offers collision-free motion dynamics inference, denoted as $f$, to predict the agent's next state $x_{t+1}$.
We denote the error in the prediction generated by the underlying RVO motion model as $q$.
Additionally, the observations of our framework or tracker can be represented by a function $h$ that projects the state $x_{t}$ to an observed state, denoted as $y_{t}$. Moreover, we denote the error between the observed states and the ground truth as $r$. We can now phrase them formally in terms of a standard particle filter as below:

\begin{equation}
x_{k+1} = f(x_{k}) + q,
\end{equation}
\begin{equation}
y_{k} = h(x_{k}) + r.
\end{equation}

\noindent In our formulation, we use RVO to infer dynamic transition, $p(x_{t}|x_{t-−1})$, for particle filtering.

\textbf{Confidence Estimation Level:} We analyze the confidence of our tracker given the number of particles. We use two metrics: propagation reliability and motion model reliability.



\paragraph{Propagation Reliability}
This is a measure of how well does the object matches the target candidate during each frame. Once the normalized weights of all the particles in the our algorithm fall below a certain threshold, we remove those particles and reduce the total number of particles used for that pedestrian. The particles are expected to have higher weights at locations that correspond to the actual positions of the tracked objects. If the number of total particles becomes less than a threshold, $N$, we resample the particles for that pedestrian and make sure that each pedestrian is approximated  by at least $N$ particles.

%

\paragraph{Motion Model Reliability}
This is a key metric in our confidence estimation. We calculate the normalized difference between the tracked state in our particle filtering framework, $t_{PF}$, and the predicted motion model state $t_{RVO}$. If this difference, $d$, is high, more particles are introduced in the system and resampled. Otherwise, we gradually reduce the number of particles and retest the confidence at each timestep. The computations related to each particle are independent and have the same overhead, and hence as a general rule of thumb, computation cost is directly proportional to the number of particles used. \textit{(see Algorithm 2, Fig. ~\ref{fig:5})}

\begin{figure}[h]
	\centering
\includegraphics[width=0.5\textwidth]{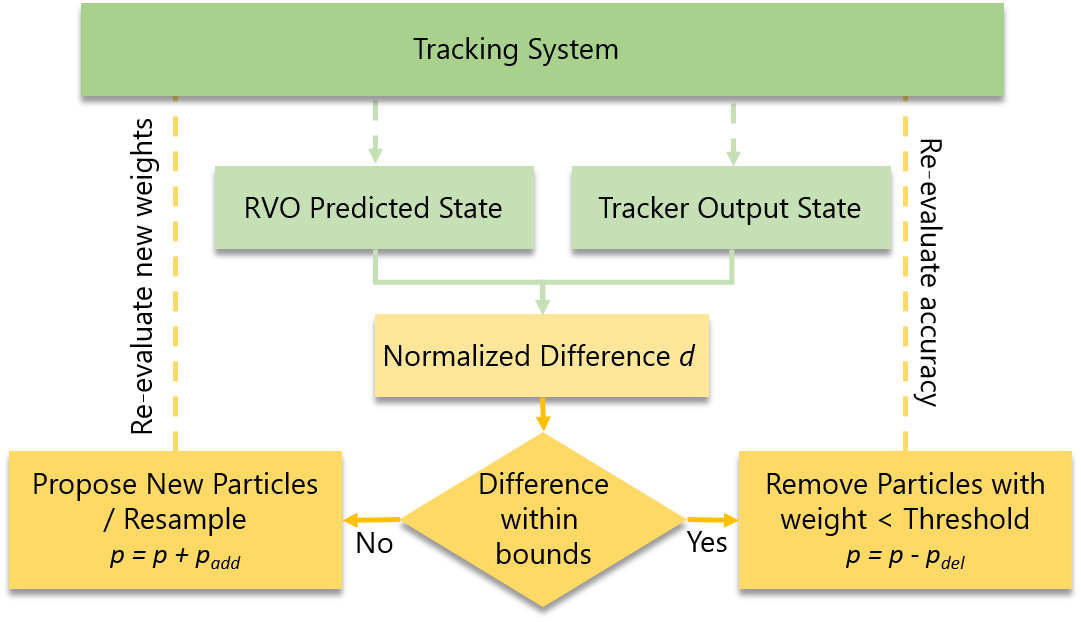}
\caption{{\em Motion Model Reliability: We highlight the motion model metric computation of the confidence estimation algorithm based on the information from the trained RVO and the k-particle filter. (See Algorithm 2)}}
	\label{fig:5}
\end{figure}

\begin{algorithm}
 \caption{Motion Model Reliability}
\emph{$d =  \|\| t_{PF} - t_{RVO} \|\|$}\;
\emph{$f = $Number of frames}\;
\emph{$p = $Number of particles at time $t$}\;
\emph{$p_{add} = $Number of additional particles introduced at time $t$}\;
\emph{$p_{del} = $Number of particles removed at time $t$}\;
\BlankLine
\For{$i\leftarrow 1$ \KwTo $f$}{
 \If{$d > $ user-set threshold}{
   $p = p + p_{add}$\;
   Resample\;
   }
	\Else
	{
	Calculate $p_{del}$ particles with the lowest weights\;
	$p = p - p_{del}$
	}
 }
\end{algorithm}

\textbf{High-Level Motion Model Level:} In this level, we learn the RVO parameters and refine our motion model framework to better match the behavior of each pedestrian. The system computes and predicts the agent trajectories using statistical inferencing techniques from the noisy tracker data.

\begin{figure}[h]
	\centering
		\includegraphics[width=0.5\textwidth]{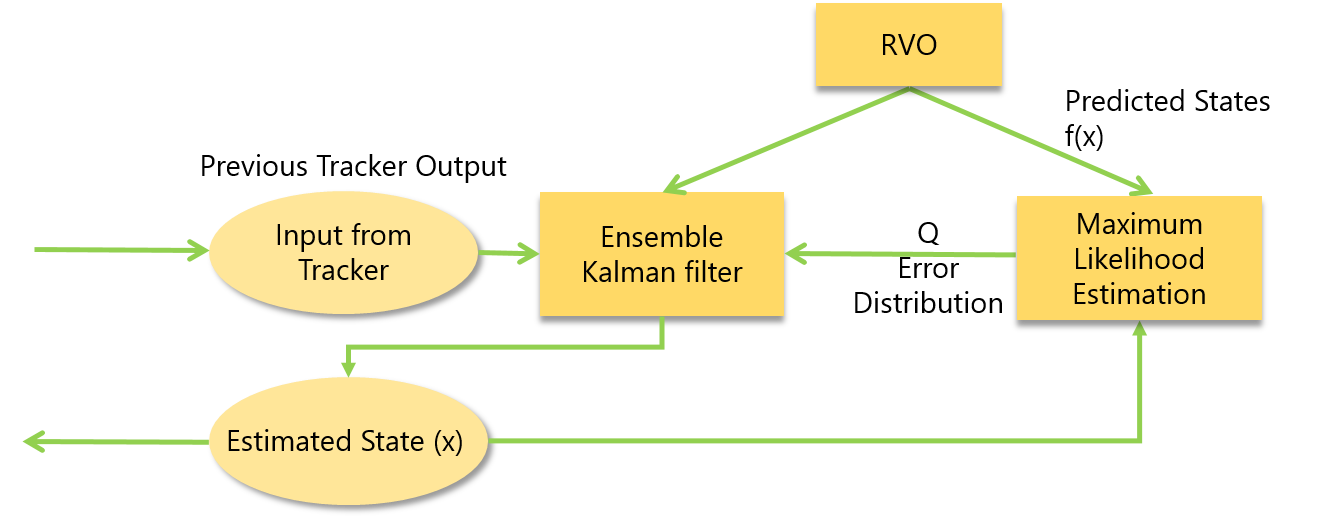}
	\caption{{\em Overview of the High-Level Motion Model Augmentation Level. This level draws input from the Confidence Estimation level, learns model parameters and improves tracking by providing feedback. The feedback is bidirectional and the model is re-trained after a 
fixed number of frames.}}
	\label{fig:6}
\end{figure}

\noindent We highlight the use of a motion model in Fig. 6 to compute the trajectory of each moving pedestrian and adaptively learn the simulation parameters based on tracked data. The resulting motion for each agent is computed using statistical techniques, which include combining Ensemble Kalman filter (EnKF) and maximum likelihood estimation algorithm to learn individual motion parameters~\cite{kim2013predicting}

\noindent The current pedestrian state is computed via a recursive process. The output of our tracker is used to recursively re-estimate the current state. The model combines the EM (Expectation-Maximization) algorithm with an ensemble Kalman Filtering approach to iteratively approximate the motion model state of each agent at every timestep. 

\noindent We perform Bayesian learning for each pedestrian. Every pedestrian can be represented by a motion model state vector $\mathbf{x}$.
Given a pedestrian's state (position, velocity and preferred velocity), we use the motion model $f$, to predict the pedestrian's next 
state $\mathbf{x}_{k+1}$. We denote the error that the motion model has in terms of predicting the state as $\mathbf{q}$ and it follows a Gaussian distribution with covariance $\mathbf{E}$. Hence,

\begin{equation} \label{eqn:pred}
\mathbf{x}_{k+1} = f(\mathbf{x}_{k}) + \mathbf{q}.
\end{equation}

Additionally, we assume that our output from the tracking stage can be represented by a function
$h$ that projects the predicted state $x_{k}$
to an observed state $z_{k}$. $r$ is the error between the observed state and the ground truth. Hence:

\begin{equation} \label{eqn:pred}
\mathbf{z}_{k} = h(\mathbf{x}_{k}) + \mathbf{r}.
\end{equation}

\noindent Our motion model level uses RVO to represent the function $f$ and EnKF to estimate the simulation parameters
which best fit the observed data. EM-algorithm is used to estimate
the model error for each pedestrian. Better estimation of the model error improves the Kalman Filtering
process, which in turn improves the pedestrian state prediction. We perform EnKF and EM steps for each
pedestrian, separately, but taking into account all the nearby pedestrians that are used in RVO motion model computation $f(x)$. This improves the accuracy of our predictor and overall trajectory computation in dense scenes or scenes with cross-flow pedestrian motion.

\begin{figure}[h]
	\centering
		\includegraphics[width=0.49\textwidth]{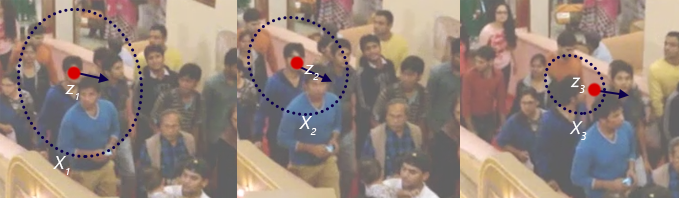}
	\caption{{\em Adaptive Refinement and Prediction. As new data is observed (red dot), we re-estimate the distribution of likely values of the RVO states (shown as a dashed ellipse). $z_{0} ...z_{t}$ are the set of observations for each pedestrian at time $t$, $x_{0} ...x_{t}$ correspond to the predicted RVO state that best reproduce the actual trajectory. The blue arrow indicates the predicted velocity vector.}}
	\label{fig:6}
\end{figure}

\begin{figure*}[!htb]
	\centering
		\includegraphics[width=1\textwidth]{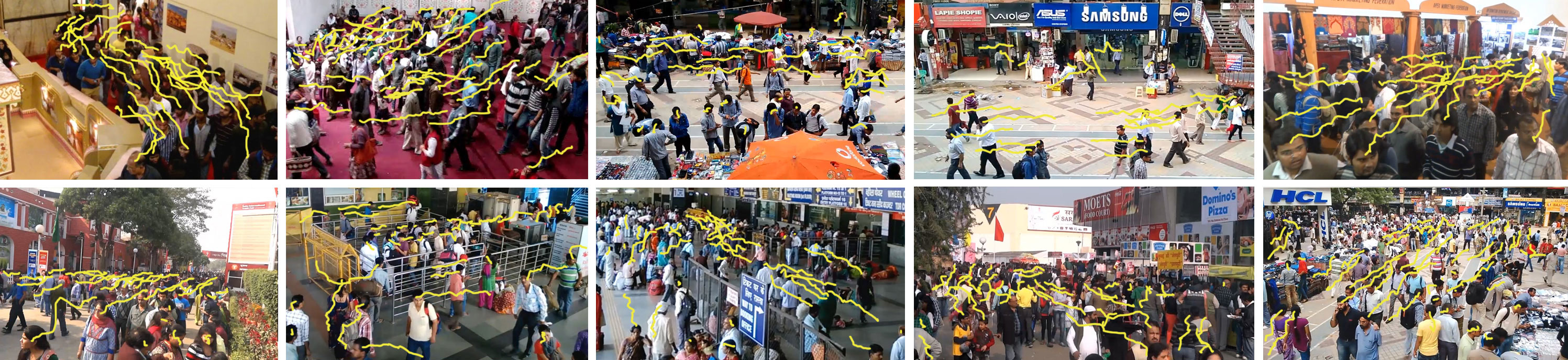}
		\label{fig:9}
		\caption{The results of our approach on the different datasets. From top-right, clockwise - \textit{IITF-1, IITF-2, NPLACE-1, NPLACE-2, IITF-3, NPLACE-3, IITF-4, NDLS-1, NDLS-2, IITF-5}. These datasets are available on our website - \textit{http://gamma.cs.unc.edu/RCrowdT}}
\vspace*{-0.13in}
\end{figure*}

\begin{table*}[ht]
{\small
\hfill{}
\scalebox{0.8}{
{\renewcommand{\arraystretch}{1.3}
 \begin{tabular}{|c|c|c|c|c|c|c|c|c|c|c|c|c|c|c|c|c|c|c|c|c|}
\hline
\multirow{2}{*}{}     & \multicolumn{2}{|c}{\textbf{IITF-1}} & \multicolumn{2}{|c}{\textbf{IITF-2}} & \multicolumn{2}{|c}{\textbf{NPLACE-1}} & \multicolumn{2}{|c}{\textbf{NPLACE-2}} & \multicolumn{2}{|c}{\textbf{IITF-3}} & \multicolumn{2}{|c}{\textbf{NPLACE-3}} & \multicolumn{2}{|c}{\textbf{IITF-4}} & \multicolumn{2}{|c}{\textbf{NDLS-1}} & \multicolumn{2}{|c}{\textbf{NDLS-2}} & \multicolumn{2}{|c|}{\textbf{IITF-5}} \\ \cline{2-21} 
                      & Acc          & FPS         & Acc          & FPS         & Acc          & FPS         & Acc          & FPS         & Acc         & FPS         & Acc         & FPS         & Acc          & FPS         & Acc          & FPS          & Acc           &  FPS          & Acc          & FPS          \\ \hline
Online Boosting       &	74\%	&	7	&	46\%	&	6	&	74\%	&	8	&	76\%	&	11	&	57\%	&	6	&	67\%	&	7	&	56\%	&	6	&	58\%	&	6	&	75\%	&	7	&	66\%	&	7	\\ \hline
KMS                   &	28\%	&	31	&	17\%	&	32	&	26\%	&	29	&	31\%	&	33	&	23\%	&	29	&	27\%	&	26	&	23\%	&	31	&	25\%	&	29	&	34\%	&	30	&	26\%	&	29	\\ \hline
SMS                   &	68\%	&	14	&	38\%	&	13	&	64\%	&	15	&	66\%	&	19	&	48\%	&	14	&	59\%	&	13	&	49\%	&	15	&	51\%	&	14	&	63\%	&	16	&	55\%	&	13	\\ \hline
ASLA                  &	72\%	&	7	&	39\%	&	7	&	70\%	&	8	&	70\%	&	12	&	51\%	&	6	&	60\%	&	7	&	50\%	&	6	&	49\%	&	7	&	68\%	&	8	&	60\%	&	8	\\ \hline
Frag                  &	41\%	&	20	&	34\%	&	21	&	40\%	&	19	&	57\%	&	19	&	54\%	&	18	&	51\%	&	21	&	48\%	&	20	&	50\%	&	22	&	66\%	&	18	&	51\%	&	19	\\ \hline
MLPF-LIN             &	63\%	&	27	&	36\%	&	26	&	67\%	&	27	&	69\%	&	28	&	51\%	&	26	&	60\%	&	28	&	52\%	&	26	&	53\%	&	26	&	68\%	&	27	&	59\%	&	27	\\ \hline
SLPF-LIN              &	64\%	&	12	&	38\%	&	12	&	68\%	&	10	&	69\%	&	12	&	51\%	&	11	&	61\%	&	11	&	53\%	&	10	&	53\%	&	9		&	68\%	&	11	&	60\%	&	12	\\ \hline
SLPF-RVO              &	71\%	&	11	&	42\%	&	10	&	73\%	&	10	&	74\%	&	11	&	53\%	&	11	&	66\%	&	11	&	53\%	&	10	&	58\%	&	10	&	72\%	&	13	&	65\%	&	11	\\ \hline

\textbf{MLPF-RVO} & \textbf{69\%}         & \textbf{27}        & \textbf{42\%}         & \textbf{26}        & \textbf{71\%}         & \textbf{26}        & \textbf{73\%}         & \textbf{28}        & \textbf{53\%}         & \textbf{26}        & \textbf{64\%}         & \textbf{27}        & \textbf{53\%}         & \textbf{26}        & \textbf{57\%}         & \textbf{26}         & \textbf{72\%}         & \textbf{27}         & \textbf{64\%}          & \textbf{27}         \\ \hline
\end{tabular}

}}}
\hfill{}
		\caption{{\em We compare the accuracy (in terms of pedestrians tracked in the video sequence) and speed (in terms of frames per second) of the following online algorithms- Online Boosting~\cite{babenko2009visual}, KMS~\cite{comaniciu2003kernel}, SMS~\cite{collins2003mean}, ASLA~\cite{jia2012visual}, Frag~\cite{adam2006robust} and also with MLPF-RVO, SLPF-RVO, MLPF-LIN and SLPF-LIN. (Abbreviation: Acc-Tracking Accuracy, FPS- Average frames per sec)}}
\label{tb:tablename}
\end{table*}

\begin{table}[H]
\caption{{\em Crowd Scenes used as Benchmarks. We highlight many attributes of crowd these videos along with density and the number of number of pedestrians tracked. We use the following abbreviations about the underlying scene: Background Variations(BV), Partial Occlusion(PO), Complete Occlusion(CO), Illumination Changes(IC) }}
\scalebox{0.8}{
    \begin{tabular}{|l|l|l|l|}
    \hline
		\textbf{Dataset} & \textbf{Challenges} & \textbf{Density} & \textbf{Pedestrians tracked}                                       \\ \hline
    NDLS-1 & BV, PO, IC & High  & 131                                     \\ \hline
    NDLS-2   & BV, PO, IC, CO & Medium & 72 \\ \hline
    NPLACE-1    & BV, PO, IC  & Medium   & 79             \\ \hline
	  NPLACE-2 & BV, PO & Low & 56                                       \\ \hline
    NPLACE-3   & BV, PO, IC, CO & High & 144\\ \hline
    IITF-1    & BV, PO, IC, CO  & High & 167\\ \hline
    IITF-2    & BV, PO, IC, CO  & High & 68                \\ \hline
    IITF-3    & BV, PO, IC, CO  & High & 189\\ \hline
    IITF-4    & BV, PO, IC, CO  & High &  116              \\ \hline
    IITF-5    & BV, PO, IC, CO  & High & 71\\ \hline
    \end{tabular}
		}

\end{table}

\noindent We use this trained motion model in our particle filter for dynamic transition and the predicted RVO state for calculating the confidence in the `Confidence Estimation' level of the algorithm.

\section{Implementation and Results}

\noindent In this section, we highlight the performance of our algorithm on different benchmarks and compare the performance with some prior techniques. (See Table II, Fig. 8)

\noindent We tested these algorithms on an Intel\textcopyright  x86 Processor (8 Cores). 8MB Cache, 3.90 GHz. Our algorithm is implemented in C++, and many components use OpenMP for exploiting multiple cores.

For our experiment we have divided our system into two phases:
\emph{Training:} This is the `motion model' level of our algorithm shown in Fig. 3. We run our input video for $k$ frames and estimate the RVO parameters.
\emph{Predict:} After training, we use the predicted state and the trained motion model for improving accuracy and for confidence calculation.

\noindent For our k-particle filter, we vary $k$ in the following manner: if there is a loss in confidence, we increase $k$ in multiples of 100. After every subsequent increase, we keep it constant for 10 frames, unless the confidence drops below our threshold. After we achieve a stable confidence estimate, we gradually decrease the number of particles by removing particles with low weights. Please refer to Fig. 9.

\begin{figure}[h]
	\centering
		\includegraphics[width=0.45\textwidth]{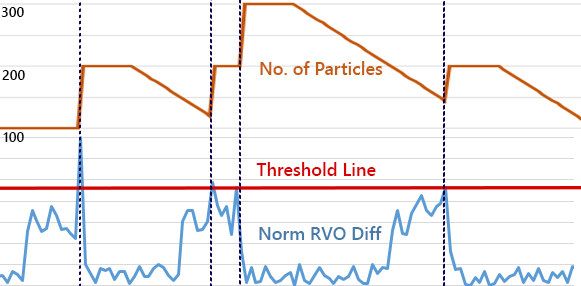}
	\caption{{\em k-particle filter implementation. The blue graph below denotes the normalized difference metric, $d$. Once $d$ exceeds a certain threshold (red line), we increase the number of particles, $p$ (denoted by orange line) by 100. After a while, we start decreasing the particles slowly to see if we are able to maintain the required confidence. This process is repeated for every pedestrian. X-Axis represents number of active particles.
}}
	\label{fig:6}
\end{figure}

We compare our approach with many well-known online tracking algorithms (as shown in Table 1). In order to demonstrate the benefits of our multi-level tracker and the improved motion model, we consider the following four combinations:

\begin{itemize}[noitemsep,nolistsep]

\item {SLPF-LIN:} In this case, the particle filter uses a constant number of particles along with constant velocity motion model.

\item {SLPF-RVO:} This uses a constant number of particles along with RVO as the motion model.

\item {MLPF-LIN:} We dynamically change the number of particles along with constant velociy.

\item{MLPF-RVO:} This uses an adaptive particle filter along with RVOs. In our benchmarks, this version achieves realtime performance with the best accuracy.

\end{itemize}

\section{Limitations, Conclusions, and Future Work}
\noindent We present  a realtime algorithm for pedestrian tracking in crowded scenes. provides a good balance between accuracy and speed. We highlight its performance on many pedestrian
datasets and can track crowded scenes in realtime on a PC with a multi-core CPU. As compared to prior algorithms of similar accuracy, we obtain 4-5 times speedup.

\noindent Our approach has some limitations related to our motion model. RVOs do not take into account physiological and psychological pedestrian traits. All pedestrians are modeled with the same sensitivity towards gender, density and and doesn't take into account heterogeneous characteristics. These may have introduced additional errors in our confidence estimation. In practice, the performance of the algorithm can vary based on various other attributes of the input video.

\noindent For future work, use improved motion models that exploit `fundamental diagrams'~\cite{curtispedestrian}, which can result in improved prediction in highly dense scenarios. In terms of performance, we would like to exploit the GPU capabilities and evaluate the performance on mobile devices.

\newcommand{\BIBdecl}{\setlength{\itemsep}{0.1 em}}
\bibliographystyle{IEEEtran}
\bibliography{IEEEabrv,root}
\end{document}